\documentclass[10pt,twocolumn,letterpaper]{article}

\usepackage[pagenumbers]{cvpr} 

\makeatletter
\@namedef{ver@everyshi.sty}{}
\makeatother
\usepackage{tikz}

\usepackage{graphicx}
\usepackage{amsmath}
\usepackage{amssymb}
\usepackage{booktabs}

\usepackage{xcolor}
\usepackage[pagebackref,breaklinks,colorlinks,bookmarks=false,urlcolor=black,citecolor={green!60!black}]{hyperref}

\usepackage{algorithm}
\usepackage{algorithmic}
\usepackage{colortbl}
\definecolor{dkgreen}{rgb}{0,0.6,0}
\definecolor{gray}{rgb}{0.5,0.5,0.5}
\definecolor{mauve}{rgb}{0.58,0,0.82}
\definecolor{LightCyan}{rgb}{0.88,1,1}

\usepackage{color}

\usepackage{tikz}
\usepackage{comment}
\usepackage{booktabs}
\usepackage{epsfig}
\usepackage{multirow}
\usepackage{enumitem}
\usepackage{bbm}
\usepackage{arydshln}

\usepackage{dsfont}
\usepackage{bm}
\usepackage{mathtools}

\usepackage{listings}
\usepackage{color}

\usepackage[pagebackref,breaklinks,colorlinks]{hyperref}

\usepackage{txfonts}
\newcommand{\heart}{\ensuremath\varheartsuit}

\usepackage[capitalize]{cleveref}
\crefname{section}{Sec.}{Secs.}
\Crefname{section}{Section}{Sections}
\Crefname{table}{Table}{Tables}
\crefname{table}{Tab.}{Tabs.}

\def\cvprPaperID{5780} 
\def\confName{CVPR}
\def\confYear{2023}

\definecolor{dkgreen}{rgb}{0,0.6,0}
\definecolor{gray}{rgb}{0.5,0.5,0.5}
\definecolor{mauve}{rgb}{0.58,0,0.82}

\makeatletter
\def\ps@myheadings{%
    \let\@oddfoot\@empty\let\@evenfoot\@empty
    \def\@evenhead{\thepage\hfil\slshape\leftmark}%
    \def\@oddhead{{\slshape\rightmark}\hfil\thepage}%
    \let\@mkboth\@gobbletwo
    \let\sectionmark\@gobble
    \let\subsectionmark\@gobble
    }
  \if@titlepage
  \renewcommand\maketitle{\begin{titlepage}%
  \let\footnotesize\small
  \let\footnoterule\relax
  \let \footnote \thanks
  \null\vfil
  \vskip 60\p@
  \begin{center}%
    {\LARGE \@title \par}%
    \vskip 3em%
    {\large
     \lineskip .75em%
      \begin{tabular}[t]{c}%
        \@author
      \end{tabular}\par}%
      \vskip 1.5em%
    {\large \@date \par}
  \end{center}\par
  \@thanks
  \vfil\null
  \end{titlepage}%
  \setcounter{footnote}{0}%
}
\else
\renewcommand\maketitle{\par
  \begingroup
    \renewcommand\thefootnote{\@fnsymbol\c@footnote}%
    \def\@makefnmark{\rlap{\@textsuperscript{\normalfont\color{black}\@thefnmark}}}%
    \long\def\@makefntext##1{\parindent 1em\noindent
            \hb@xt@1.8em{%
                \hss\@textsuperscript{\normalfont\@thefnmark}}##1}%
    \if@twocolumn
      \ifnum \col@number=\@ne
        \@maketitle
      \else
        \twocolumn[\@maketitle]%
      \fi
    \else
      \newpage
      \global\@topnum\z@   
      \@maketitle
    \fi
    \thispagestyle{plain}\@thanks
  \endgroup
  \setcounter{footnote}{0}%
}
\makeatother

\makeatletter
\newcommand\fs@nobottomruled{\def\@fs@cfont{\bfseries}\let\@fs@capt\floatc@ruled
  \def\@fs@pre{}
  \def\@fs@post{}
  \def\@fs@mid{\kern2pt\hrule\kern2pt}%
  \let\@fs@iftopcapt\iftrue}
\makeatother

\makeatletter
\DeclareRobustCommand\bmvaOneDot{\futurelet\@let@token\bmv@onedotaux}
\def\bmv@onedotaux{\ifx\@let@token.\else.\null\fi\xspace}
\makeatother
\def\eg{\emph{e.g}\bmvaOneDot}

\def\etc{\emph{etc}\bmvaOneDot}
\def\ie{\emph{i.e}\bmvaOneDot}

\def\wrt{\emph{w.r.t}\bmvaOneDot}

\renewcommand\vec[1]{\ensuremath\boldsymbol{#1}}
\renewcommand\cdots{...}

\newcommand{\tM}{\vec{\mathcal{M}}}

\newcommand{\mbrp}[1]{\mathbb{R}_{+}^{#1}}
\newcommand{\mbr}[1]{\mathbb{R}^{#1}}

\newcommand{\idx}[1]{\mathcal{I}_{#1}}

\def\eg{\emph{e.g.}}

\newcommand{\mP}{\boldsymbol{\Theta}}

\newcommand{\stkout}[1]{{\ifmmode\text{\sout{\ensuremath{#1}}}\else\sout{#1}\fi}}

\newcommand{\lei}{\textcolor{black}}
\newcommand{\leicr}{\textcolor{black}}
\begin{document}

\title{\lei{3Mformer: } Multi-order Multi-mode Transformer \\for Skeletal Action Recognition}


\author{Lei Wang\textsuperscript{$\dagger,\S$}\qquad Piotr Koniusz\thanks{Corresponding author. This paper is accepted by CVPR 2023.}\textsuperscript{$\;\;,\S,\dagger$}\\
$^{\dagger}$Australian National University, $^\S$Data61\heart CSIRO\\
$^\S$firstname.lastname@data61.csiro.au
}

\maketitle

\begin{abstract}
Many  skeletal action recognition models use GCNs to represent the human body by 3D body joints connected body parts. 
GCNs aggregate \lei{one- or few-hop} graph neighbourhoods, 
and ignore the dependency between  not linked  body joints. 
We propose to form hypergraph to  model hyper-edges between graph nodes (\eg, third- and fourth-order hyper-edges capture three and four nodes) which help capture higher-order motion patterns of groups of body joints. 
We split action sequences into temporal blocks, Higher-order Transformer (HoT) produces embeddings of each temporal block based on (i) the body joints, 
(ii) pairwise links of body joints and (iii) higher-order hyper-edges of skeleton body joints. 
%
%
We  combine such HoT embeddings of hyper-edges of orders $1,\cdots,r$ by a novel \lei{Multi-order Multi-mode Transformer (3Mformer)} with two modules whose order can be exchanged to achieve \lei{{coupled-mode}} attention on {coupled-mode} tokens based on `channel-temporal block', `\lei{order-}channel-body joint', `channel-hyper-edge (any order)' and `channel-only' pairs. The first module, called Multi-order Pooling (MP)\lei{,} additionally learns weighted aggregation along the hyper-edge mode, whereas the second module, Temporal block Pooling (TP)\lei{,} aggregates along the temporal block\footnote{For brevity, we write  $\tau$ temporal blocks per sequence but $\tau$ varies.} mode.
Our end-to-end trainable network yields state-of-the-art results compared to  GCN-, transformer- and   hypergraph-based counterparts. 
\end{abstract}

\section{Introduction}

Action Recognition has  applications in video surveillance, human-computer interaction, sports analysis, and virtual reality~\cite{lei_thesis_2017,lei_icip_2019, lei_tip_2019, Wang_2019_ICCV, kon_tpami2020b, koniusz2021high, 10.1145/3474085.3475572, wang20213d, qin_tnnls_22, udtw_eccv22, Wang_2022_ACCV}. 
%
\lei{Different from video-based methods which mainly focus on modeling the spatio-temporal representations from RGB frames and/or optical flow~\cite{lei_thesis_2017,lei_icip_2019,lei_tip_2019, Wang_2019_ICCV, 10.1145/3474085.3475572, koniusz2021high},  skeleton sequences, representing a spatio-temporal evolution of 3D body joints, have been proven  robust against sensor noises and  effective in action recognition while being computationally and storage efficient~\cite{lei_thesis_2017, lei_tip_2019, kon_tpami2020b, wang20213d, qin_tnnls_22, udtw_eccv22, Wang_2022_ACCV}.}
The skeleton data is usually obtained by either localization of 2D/3D coordinates of human body joints with the depth sensors or pose estimation algorithms applied to videos~\cite{Cao_2017_CVPR}. Skeleton sequences enjoy (i) simple structural connectivity of skeletal graph and (ii) temporal continuity of 3D body joints evolving in time. While temporal evolution of each body joint is highly informative, embeddings of separate body joints are insensitive to relations between  body parts. Moreover, while the links between adjacent 3D body joints (following the structural connectivity) are very informative as they model relations, these links represent highly correlated nodes in the sense of their temporal evolution. Thus, modeling larger groups of 3D body joints as hyper-edges can capture more complex spatio-temporal motion dynamics.

The existing graph-based models mainly differ by how they handle temporal information.  
\lei{Graph Neural Network (GNN) may encode spatial neighborhood of the node followed by aggregation by LSTM~\cite{Si_2019_CVPR, 8784712}. 
Alternatively, Graph Convolutional Network (GCN) may perform spatio-temporal convolution in the neighborhood of each node~\cite{stgcn2018aaai}.} 
Spatial GCNs perform convolution within one or two hop distance of each node, \eg, spatio-temporal GCN model called ST-GCN \cite{stgcn2018aaai} models spatio-temporal vicinity of each 3D body joint. 
%
As ST-GCN applies convolution along structural connections (links between body joints), structurally distant joints, which may cover key patterns of actions, are largely ignored. 
ST-GCN captures ever larger neighborhoods as layers are added but suffers from  oversmoothing that can be mitigated by linear GCNs \cite{ssgc_hao,coles_hao,glen_hao}. 
%

Human actions are  associated with interaction groups of skeletal joints, \eg, wrist alone, head-wrist, head-wrist-ankles, \etc. The impact of these groups of joints on each action differs, and the degree of influence of each joint should be learned. Accordingly, designing a better model for skeleton data is vital given the topology of skeleton graph  is  suboptimal. 
While GCN can be applied to a fully-connected graph (\ie, 3D body joints as densely connected graph nodes), Higher-order Transformer (HoT) \cite{kim2021transformers} has been proven more efficient.

Thus, we propose to use hypergraphs with hyper-edges of order $1$ to $r$ to effectively represent skeleton data for action recognition. 
Compared to  GCNs, 
our encoder contains an MLP followed by three HoT branches that encode first-, second- and higher-order hyper-edges, \ie, set of body joints, edges between pairs of nodes, hyper-edges between triplets of nodes, \etc. Each branch  has its own learnable parameters, and processes  temporal blocks\footnote{Each temporal block enjoys a locally factored out (removed) temporal mode, which makes each block representation  compact.} one-by-one.

We notice that (i) the number of hyper-edges of $J$ joints grows rapidly with order $r$, \ie, $\binom{J}{i}$ for $i=1,\cdots,r$,  embeddings of the highest order 
dominate lower orders in terms of volume if such embeddings are merely concatenated, and (ii) long-range temporal dependencies of feature maps are insufficiently explored, as  sequences are split into $\tau$ temporal blocks  for computational tractability. 

Merely concatenating outputs of  HoT branches of orders $1$ to $r$, and across $\tau$ blocks, is  sub-optimal.  
Thus, our \lei{Multi-order Multi-mode Transformer (3Mformer)} with two modules whose order can be exchanged, realizes a variation of coupled-mode tokens based on `channel-temporal block', `\lei{order-}channel-body joint', `channel-hyper-edge (any order)' and `channel-only' pairs. As HoT operates  block-by-block, `channel-temporal block' tokens and weighted hyper-edge aggregation in Multi-order Pooling (MP) help combine information flow block-wise. Various {coupled-mode} tokens help improve results further due to different focus of each attention mechanism. As the {block-temporal} mode needs to be aggregated (number of blocks varies across sequences), Temporal block Pooling (TP) can use rank pooling~\cite{10.1109/TPAMI.2016.2558148}, second-order~\cite{tsungyu_eccv2018,Gao_2019_CVPR,NEURIPS2018_17c276c8, Girdhar_17b_AttentionalPoolingAction,zhang2020sopaccv,kon_tpami2020a,simon_cvpr2023} or higher-order pooling~\cite{7926605, koniusz2021high,kon_tpami2020b,zhang2022kernelized,zhang2022time}. 

\vspace{0.2cm}
In summary, our main contributions are listed as follows:$\!\!\!$

\renewcommand{\labelenumi}{\roman{enumi}.}
\begin{enumerate}[leftmargin=0.6cm]
\item We model the skeleton data as hypergraph of orders $1$ to $r$ (set, graph and/or hypergraph), where human body joints serve as nodes. Higher-order Transformer embeddings of such formed hyper-edges represent various groups of 3D body joints and capture various higher-order dynamics important for action recognition.
\item As HoT embeddings represent individual hyper-edge order and block, we introduce a novel Multi-order Multi-mode Transformer (3Mformer) with two modules, Multi-order Pooling and Temporal block Pooling. Their goal is to form coupled-mode tokens such as `channel-temporal block', `order-channel-body joint', `channel-hyper-edge (any order)' and `channel-only', and perform weighted hyper-edge aggregation and temporal block aggregation.
%
\end{enumerate}

Our 3Mformer outperforms  other GCN- and hypergraph-based models on NTU-60, NTU-120, Kinetics-Skeleton and Northwestern-UCLA by a large margin.



\begin{figure*}[t]
\centering\includegraphics[width=1\linewidth]{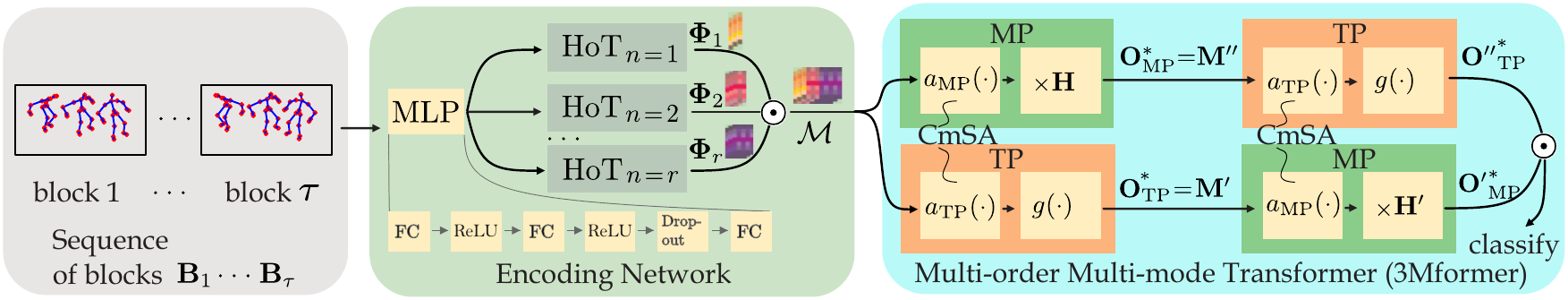}
\caption{Pipeline overview. Each sequence is split into $\tau$ temporal blocks $\mathbf{B}_1,\cdots,\mathbf{B}_\tau$. Subsequently, each block is embedded by a simple MLP into $\mathbf{X}_1,\cdots,\mathbf{X}_\tau$, which are passed to  Higher-order Transformers (\lei{HoT} ($n\!=\!1,\cdots,r$)) in order to obtain feature tensors $\mathbf{\Phi}_1,\cdots,\mathbf{\Phi}_\tau$. These tensors are subsequently concatenated by $\odot$ along the hyper-edge mode into a multi-order feature tensor $\boldsymbol{\mathcal{M}}$. The final step is a \lei{Multi-order Multi-mode Transformer (3Mformer} from Section \ref{sec:appr}), which contains two \lei{complementary} branches, MP$\rightarrow$TP and TP$\rightarrow$MP, whose outputs are concatenated by $\odot$ and passed to the classifier. MP and TP perform the \lei{\leicr{Coupled-mode}} Self-Attention (\leicr{CmSA}) with the so-called \leicr{coupled-mode} tokens,  based
on `channel-temporal block’, `\lei{order-}channel-body joint’, `channel-hyper-edge’ and `channel-only’ pairs. To this end, MP contains also weighted pooling along hyper-edge mode by learnable matrix $\mathbf{H}$ (and $\mathbf{H}'$ in another branch). TP contains also \lei{block-temporal} pooling denoted by $g(\cdot)$ whose role is to capture \lei{block-temporal} order with average, maximum, rank pooling, \etc. In our experiments we show that such designed MP and TP are able to efficiently process hyper-edge feature representations from HoT branches. Appendix \ref{app:3mf} shows full visualization of our 3Mformer.}
\label{fig:pipeline}
\end{figure*}

\section{Related Work}
Below we describe popular action recognition models for skeletal data.

\vspace{0.1cm}
\noindent{\bf Graph-based models}. Popular GCN-based models include the Attention enhanced Graph Convolutional LSTM network (AGC-LSTM)~\cite{Si_2019_CVPR}, the Actional-Structural GCN (AS-GCN)~\cite{Li_2019_CVPR}, Dynamic Directed GCN (DDGCN)~\cite{10.1007/978-3-030-58565-5_45}, Decoupling GCN with DropGraph module~\cite{10.1007/978-3-030-58586-0_32}, Shift-GCN~\cite{cheng2020shiftgcn}, Semantics-Guided Neural Networks (SGN)~\cite{Zhang_2020_CVPR2}, AdaSGN~\cite{Shi_2021_ICCV}, Context Aware GCN (CA-GCN)~\cite{Zhang_2020_CVPR1}, Channel-wise Topology Refinement Graph Convolution Network (CTR-GCN)~\cite{chen2021channel} and a family of Efficient GCN (EfficientGCN-Bx)~\cite{9729609}.
Although  GCN-based models enjoy good performance, they have shortcomings, \eg, convolution and/or pooling are applied over \lei{one- or few-hop neighborhoods, \eg, ST-GCN~\cite{stgcn2018aaai}}, according to the human skeleton graph (body joints linked up according to connectivity of human body parts). Thus,  indirect links between various 3D body joints such as hands and legs are ignored. 
In contrast, our model is not restricted by the structure of typical human body skeletal graph. Instead, 3D body joints are nodes which form hyper-edges of orders $1$ to $r$.

\vspace{0.1cm}
\noindent{\bf Hypergraph-based models}. Pioneering work on capturing groups of nodes across time uses tensors \cite{kon_tpami2020b} to represent the 3D human body joints to exploit the kinematic relations among the adjacent and non-adjacent joints. Representing the human body as a hypergraph is adopted in~\cite{ijcai2020-109} via a semi-dynamic hypergraph neural network that captures richer information than GCN. A hypergraph GNN \cite{9329123}  captures both spatio-temporal information and higher-order dependencies for skeleton-based action recognition. Our work is somewhat closely related to these works, but we jointly use hypergraphs of order $1$ to $r$ to obtain rich hyper-edge embeddings based on Higher-order Transformers.

\vspace{0.1cm}
\noindent{\bf Transformer-based models}. Action recognition with transformers includes self-supervised video transformer \cite{ranasinghe2022selfsupervised} that matches the features from different views (a popular strategy in self-supervised GCNs \cite{costa_yifei,sfa_yifei}), the end-to-end trainable Video-Audio-Text-Transformer (VATT)~\cite{akbari2021vatt} for learning multi-model representations from unlabeled raw video, audio and text through the multimodal contrastive losses, and the Temporal Transformer Network with Self-supervision (TTSN)~\cite{DBLP:journals/corr/abs-2112-07338}.  Motion-Transformer~\cite{10.1145/3444685.3446289} captures the temporal dependencies via a self-supervised pre-training on human actions, Masked Feature Prediction (MaskedFeat)~\cite{DBLP:journals/corr/abs-2112-09133} pre-trained on unlabeled videos with MViT-L learns abundant visual representations, and video-masked autoencoder (VideoMAE)~\cite{DBLP:journals/corr/abs-2203-12602} with vanilla ViT uses the masking strategy. In contrast to these works, we use three HoT branches of model \cite{kim2021transformers}, and we model hyper-edges of orders $1$ to $r$ by forming several multi-mode token variations in \lei{3Mformer}.

\vspace{0.1cm}
\noindent{\bf Attention.} In order to improve feature representations, attention captures relationship between tokens. Natural language processing and computer vision have driven recent developments in attention mechanisms based on transformers~\cite{NIPS2017_3f5ee243,dosovitskiy2021an}. 
Examples include the hierarchical Cross Attention Transformer (CAT)~\cite{DBLP:journals/corr/abs-2106-05786}, Cross-attention
by Temporal Shift with CNNs~\cite{https://doi.org/10.48550/arxiv.2204.00452}, Cross-Attention Multi-Scale Vision Transformer (CrossViT) for image classification~\cite{DBLP:journals/corr/abs-2103-14899} and Multi-Modality Cross Attention (MMCA) Network for image and sentence matching~\cite{Wei_2020_CVPR}. 
In GNNs, attention can be defined over edges~\cite{velickovic2018graph,conf/uai/ZhangSXMKY18} or over nodes~\cite{10.1145/3219819.3219980}. In this work, we use the attention with hyper-edges of several orders from HoT branches serving as tokens, and {coupled-mode} attention with {coupled-mode} tokens based on `channel-temporal block’, `{order-}channel-body joint’, `channel-hyper-edge
(any order)’ and `channel-only’ pairs formed in {3Mformer}.


\section{Background}

Below we describe  foundations necessary for our work. 

\vspace{0.1cm}
\noindent{\bf Notations.} 
\lei{$\idx{K}$ stands for the index set $\{1,2,\cdots,K\}$. Regular fonts are scalars; vectors are denoted by lowercase boldface letters, \eg, $\textbf{x}$; matrices by the uppercase boldface, \eg, {\textbf{M}}; and tensors by calligraphic letters, \eg, $\tM$. 
An $r$th-order tensor is denoted as $\tM \in \mbr{I_1 \times I_2 \times \cdots \times I_r}$,
and the mode-$m$ matricization of $\tM$ is denoted as $\tM_{(m)}\in \mbr{I_m \times (I_1 \cdots I_{m-1}  I_{m+1}  \cdots I_{r})}$.}

\vspace{0.1cm}
\noindent{\bf Transformer layers}~\cite{NIPS2017_3f5ee243,dosovitskiy2021an}. A transformer encoder layer   $f\!: \mbr{J\times d} \rightarrow \mbr{J \times d}$  consists of two sub-layers: (i) a self-attention  $a\!: \mbr{J \times d} \rightarrow \mbr{J \times d}$ and (ii) an element-wise feed-forward $\text{MLP}\!:\mbr{J \times d} \rightarrow \mbr{J \times d}$. For a set of $J$ nodes with ${\bf X} \!\in\!\mbr{J \times d}$, where ${\bf x}_i$ is a feature vector of node $i$, a transformer layer\footnote{Normalizations after $a(\cdot)$ \& MLP$(\cdot)$ are omitted for simplicity.} computes:
\begin{align}
    & a({\bf x}_i)\!=\!{\bf x}_i\!+\!\sum_{h=1}^H\sum_{j=1}^J\alpha_{ij}^h{\bf x}_j{\bf W}_h^V{\bf W}_h^O, \label{eq:transformer-attn}\\
    & f({\bf x}_i)\!=\!a({\bf x}_i)\!+\!\text{MLP}(a({\bf X}))_i, \label{eq:transformer-enc}
\end{align}
where $H$ and $d_H$ denote respectively the number of heads and the head size, ${\boldsymbol{\alpha}}^h\!=\!\sigma\big({\bf X}{\bf W}_h^Q({\bf X}{\bf W}_h^{K})^\top\big)$ is the attention coefficient, ${\bf W}_h^O\!\in\!\mbr{d_H\times d}$, and ${\bf W}_h^V$, ${\bf W}_h^K$, ${\bf W}_h^Q \!\in\!\mbr{d\times d_H}$. 


\vspace{0.1cm}
\noindent{\bf Higher-order transformer layers}~\cite{kim2021transformers}. Let the HoT layer be $f_{m\rightarrow n}\!:\mbr{J^m\times d}\!\rightarrow\!\mbr{J^n \times d}$ with two sub-layers: (i) a higher-order self-attention $a_{m\rightarrow n}\!:\mbr{J^m\times d}\!\rightarrow\!\mbr{J^n\times d}$ and (ii) a feed-forward  $\text{MLP}_{n\rightarrow n}\!:\mbr{J^n\times d}\!\rightarrow\!\mbr{J^n\times d}$. Moreover, let indexing vectors ${\bf i}\in\idx{J}^m\equiv\idx{J}\!\times\!\idx{J}\!\times\!\cdots\!\times\!\idx{J}$ ($m$ modes) and ${\bf j}\in\idx{J}^n\equiv\idx{J}\!\times\!\idx{J}\!\times\!\cdots\!\times\!\idx{J}$ ($n$ modes). For the input tensor ${\bf X}\!\in\!\mbr{J^m\times d}$ with hyper-edges of order $m$, a HoT layer evaluates:
\begin{align}
    & a_{m \rightarrow n}({\bf X})_{\boldsymbol j}\!=\!\sum_{h=1}^H\sum_\mu\sum_{\boldsymbol i}{\boldsymbol{\alpha}}_{{\boldsymbol i}, {\boldsymbol j}}^{h,\mu}{\bf X}_{\boldsymbol i}{\bf W}_{h, \mu}^V{\bf W}_{h, \mu}^O \label{eq:hot-attn}\\
    & \text{MLP}_{n\rightarrow n}(a_{m\rightarrow n}({\bf X}))\!=\!\text{L}_{n\rightarrow n}^2(\text{ReLU}(\text{L}_{n\rightarrow n}^1(a_{m\rightarrow n}({\bf X})))), \label{eq:hot-mlp}\\
    & f_{m\rightarrow n}({\bf X})\!=\!a_{m \rightarrow n}({\bf X})\!+\!\text{MLP}_{n\!\rightarrow \!n}(a_{m\rightarrow n}({\bf X})), \label{eq:hot-enc}
\end{align}
where ${\boldsymbol \alpha}^{h, \mu}\!\in\!\mbr{J^{m+n}}$ is the so-called attention coefficient tensor with multiple heads, and ${\boldsymbol \alpha}^{h, \mu}_{{\bf i},{\bf j}}\!\in\!\mbr{J}$ is a vector, ${\bf W}_{h, \mu}^V\!\in\!\mbr{d\times d_H}$ and ${\bf W}_{h, \mu}^O\!\in\!\mbr{d_H\times d}$ are learnable parameters. Moreover, $\mu$ indexes over the so-called equivalence classes of order-$(m\!+\!n)$ in the same partition of nodes,  $\text{L}_{n\rightarrow n}^1\!:\mbr{J^n\times d}\rightarrow \mbr{J^n\times d_F}$ and $\text{L}_{n\rightarrow n}^2\!:\mbr{J^n\times d_F}\rightarrow \mbr{J^n\times d}$ are equivariant linear layers and $d_F$ is the hidden dimension.

To compute each attention tensor ${\boldsymbol \alpha}^{h,\mu}\!\in\!\mbr{J^{m+n}}$ from the input tensor ${\bf X}\!\in\!\mbr{J^m\times d}$ of hyper-edges of order $m$, from the higher-order query and key, we obtain:
\begin{equation}
  {{\boldsymbol \alpha}_{{\boldsymbol i}, {\boldsymbol j}}^{h,\mu}} \!=\!
    \begin{cases}
      \frac{\sigma({\bf Q}_{\boldsymbol j}^{h,\mu}, {\bf K}_{\boldsymbol i}^{h,\mu})}{Z_{\boldsymbol j}}\;\quad({\boldsymbol i}, {\boldsymbol j}) \!\in\! \mu\\
      \quad\quad 0 \quad\quad\;\text{otherwise},
    \end{cases}
    \label{eq:att-tensor}
\end{equation}
where ${\bf Q}^\mu\!=\!\text{L}_{m\rightarrow n}^\mu({\bf X})$, ${\bf K}^\mu\!=\!\text{L}_{m\rightarrow m}^\mu({\bf X})$, and normalization constant $Z_{\boldsymbol j}\!=\!\sum_{{\boldsymbol i}:({\boldsymbol i}, {\boldsymbol j})\in \mu}\sigma({\bf Q}_{\boldsymbol j}^\mu, {\bf K}_{\boldsymbol i}^\mu)$. Finally, kernel attention  in Eq.~\eqref{eq:att-tensor} can be approximated with RKHS feature maps $\psi\in\mbrp{d_K}$ for efficacy as $d_K\ll d_H$. Specifically, we have $\sigma\big({\bf Q}_{\boldsymbol j}^{h,\mu}, {\bf K}_{\boldsymbol i}^{h,\mu}\big)\approx{\boldsymbol \psi}\big({\bf Q}_{\boldsymbol j}^{h,\mu}\big)^\top{\boldsymbol \psi}\big({\bf K}_{\boldsymbol i}^{h,\mu}\big)$ as in \cite{pmlr-v119-katharopoulos20a,choromanski2021rethinking}. 
We  choose the performer kernel~\cite{choromanski2021rethinking} due to its good  performance. 

As  query and key tensors are computed from the input tensor ${\bf X}$ using the equivariant linear layers, the transformer encoder layer $f_{m\rightarrow n}$ satisfies the permutation equivariance.

\section{Approach}
\label{sec:appr}


\lei{Skeletal Graph~\cite{stgcn2018aaai} and Skeletal Hypergraph~\cite{ijcai2020-109,9329123} are  popular for modeling edges and hyper-edges. In this work, we use the Higher-order Transformer (HoT)~\cite{kim2021transformers} as a backbone encoder.}
\subsection{\lei{Model Overview}}

Fig.~\ref{fig:pipeline} shows that our framework contains a simple 3-layer MLP unit (FC, ReLU, FC, ReLU, Dropout, FC), three HoT blocks with each HoT for each type of input (\ie, body joint feature set, graph and hypergraph of body joints), followed by Multi-order Multi-mode Transformer (3Mformer) with two modules (i) Multi-order Pooling (MP) and (ii) Temporal block Pooling (TP). 
The goal of 3Mformer is to form {coupled-mode} tokens (explained later) such as `channel-temporal block', `order-channel-body joint', `channel-hyper-edge (any order)' and `channel-only', and perform weighted hyper-edge aggregation and temporal block aggregation. Their outputs are further concatenated and passed to an FC layer for classification.

\vspace{0.1cm}
\noindent\textbf{MLP unit.} 
The MLP unit takes $T$ neighboring frames, each with $J$ 2D/3D skeleton body joints, forming one temporal block. In total, depending on stride $S$, we obtain some $\tau$ temporal blocks (a block captures the short-term temporal evolution), In contrast, the long-term temporal evolution is modeled with HoT and \lei{3Mformer}. Each temporal block is encoded by the MLP into a $d\!\times\!J$ dimensional feature map.  

\vspace{0.1cm}
\noindent\textbf{HoT branches.} 
We stack  $r$  branches of  HoT, each taking  embeddings ${\bf X}_t\in\mbr{d\!\times\!J}$ where  $t\in\idx{\tau}$ denotes a temporal block. 
HoT branches output hyper-edge feature representations of size $m\in\idx{r}$ as ${\bf\Phi}'_m\in\mbr{J^m \times d'}$ for order $m\in\idx{r}$. 

For the first-, second- and higher-order stream outputs ${\bf\Phi}'_1,\cdots,{\bf\Phi}'_r$, we (i) swap feature channel and hyper-edge modes, (ii) extract the upper triangular of tensors, and we concatenate along the block-temporal mode, so we have ${\bf\Phi}_m\in\mbr{d'\times N_{E_m}\times\tau}$, where $N_{E_m}\!=\!\binom{J}{m}$. Subsequently, we concatenate ${\bf\Phi}_1,\cdots,{\bf\Phi}_r$ along the hyper-edge mode and obtain a multi-order feature tensor $\lei{\tM}\!\in\!\mbr{d'\!\times\!N\!\times\!\tau}$ where the total number of hyper-edges across all orders is $N=\sum_{m=1}^r\binom{J}{m}$. 

\vspace{0.1cm}
\noindent\textbf{3Mformer.} Our Multi-order Multi-mode Transformer (3Mformer) with \leicr{Coupled-mode} Self-Attention (\leicr{CmSA}) is used for the fusion of information flow inside the multi-order feature tensor ${\tM}$, and finally, the output from 3Mformer is passed to a classifier for classification.

\subsection{\lei{\leicr{Coupled-mode} Self-Attention}}

\vspace{0.1cm}
\noindent\lei{\textbf{\leicr{Coupled-mode} tokens.} We are inspired by the attentive regions of the one-class token in the standard Vision Transformer (ViT)~\cite{NIPS2017_3f5ee243} that can be leveraged to form a class-agnostic localization map. We investigate if the transformer model can also effectively capture the \leicr{coupled-mode} attention for more discriminative classification tasks, \eg, tensorial skeleton-based action recognition by learning the \leicr{coupled-mode} tokens within the transformer. 
To this end, we propose a Multi-order Multi-mode Transformer (3Mformer), which uses \leicr{coupled-mode} tokens to jointly learn various higher-order motion dynamics among channel-, block-temporal-, body joint- and order-mode. Our 3Mformer can successfully produce \leicr{coupled-mode} relationships from \leicr{CmSA} mechanism corresponding to different tokens. Below we introduce our \leicr{CmSA}.
}

\lei{Given the order-$r$ tensor $\tM \in \mbr{I_1 \times I_2 \times \cdots \times I_r}$, to form the joint mode token, we perform the mode-$m$ matricization of $\tM$ to obtain $\textbf{M} \equiv \tM_{(m)}^\top \in \mbr{(I_1 \cdots I_{m-1}  I_{m+1}  \cdots I_{r}) \times I_m}$, and the \leicr{coupled-token} for $\textbf{M}$ is formed. 
For example, for a given 3rd-order tensor that has feature channel-, hyper-edge- and temporal block-mode, we can form  `channel-temporal block', `channel-hyper-edge (any order)' and `channel-only' pairs; and if the given tensor is used as input and outputs a new tensor which produces new mode, \eg, body joint-mode, we can form the `order-channel-body joint' token.
In the following sections, for simplicity, we use reshape for the matricization of tensor to form different types of \leicr{coupled-mode} tokens.  
Our \leicr{CmSA} is given as:
\begin{equation}
    \!\!a({\bf Q}, {\bf K}, {\bf V}) \!=\!\text{SoftMax}\left(\frac{{\bf Q}{\bf K}^\top}{\sqrt{d_{K}}}\right)\!{\bf V},\label{eq:jmsa}\!
\end{equation}
where $\sqrt{d_{K}}$ is the scaling factor, ${\bf Q}\!=\!{\bf W}^q{\bf M}$, ${\bf K}\!=\!{\bf W}^k{\bf M}$ and ${\bf V}\!=\!{\bf W}^v{\bf M}$ are the query, key and value, respectively, and $\textbf{M} \equiv \tM_{(m)}^\top$. Moreover, ${\bf Q}$, ${\bf K}$, ${\bf V}\!\in\!\mbr{(I_1 \cdots I_{m-1}  I_{m+1}  \cdots I_{r}) \times I_m}$ and ${\bf W}^q$, ${\bf W}^k$, ${\bf W}^v\!\in\!\mbr{(I_1 \cdots I_{m-1}  I_{m+1}  \cdots I_{r}) \times (I_1 \cdots I_{m-1}  I_{m+1}  \cdots I_{r})}$  are learnable weights. We notice that various \leicr{coupled-mode} tokens have different `focus' of attention mechanisms, and we apply them in our 3Mformer for the fusion of multi-order feature representations.
}

\subsection{\lei{Multi-order Multi-mode Transformer}}


%
Below we introduce \lei{Multi-order Multi-mode Transformer (3Mformer) with  Multi-order Pooling (MP) block and Temporal block Pooling (TP) block, which are cascaded into two branches (i) MP$\rightarrow$TP and (ii) TP$\rightarrow$MP, to achieve different types of  \leicr{coupled-mode} tokens.}

\subsubsection{\lei{Multi-order Pooling (MP) Module}}

\vspace{0.1cm}
\noindent{\bf \lei{\leicr{CmSA}} in MP.} We reshape the multi-order feature representation $\lei{\tM}\!\in\!\mbr{d'\!\times\!N\!\times\!\tau}$ into ${\bf M}\!\in\!\mbr{d'\tau\!\times\!N}$ (or reshape the output from TP explained later into ${\bf M}'\!\in\!\mbr{d'\!\times\!N}$) to let the model attend to different types of feature representations. Let us simply denote $d''\!=\!d'\tau$ (or $d''\!=\!d'$) depending on the source of input. We form an \leicr{coupled-mode} \lei{self-attention} (if $d''=d'\tau$, we have, \ie, `channel-temporal block' token\lei{; if $d''=d'$, we have `channel-only' token}):
\begin{equation}
\!\!a_\text{MP}({\bf Q}_\text{MP}, {\bf K}_\text{MP}, {\bf V}_\text{MP}) \!=\!\text{SoftMax}\left(\frac{{\bf Q}_\text{MP}{\bf K}_\text{MP}^\top}{\sqrt{d_{K_\text{MP}}}}\right)\!{\bf V}_\text{MP},\label{eq:order-attn}\!
\end{equation}
where $\sqrt{d_{K_\text{MP}}}$ is the scaling factor, ${\bf Q}_\text{MP}\!=\!{\bf W}_\text{MP}^q{\bf M}$, ${\bf K}_\text{MP}\!=\!{\bf W}_\text{MP}^k{\bf M}$ and ${\bf V}_\text{MP}\!=\!{\bf W}_\text{MP}^v{\bf M}$  (we can use here ${\bf M}$ or ${\bf M}'$) are the query, key and value. Moreover, ${\bf Q}_\text{MP}$, ${\bf K}_\text{MP}$, ${\bf V}_\text{MP}\!\in\!\mbr{d''\times N}$ and ${\bf W}_\text{MP}^q$, ${\bf W}_\text{MP}^k$, ${\bf W}_\text{MP}^v\!\in\!\mbr{d''\times d''}$  are learnable weights. Eq.~\eqref{eq:order-attn} is a self-attention layer  which reweighs ${\bf V}_\text{MP}$ based on the correlation between  ${\bf Q}_\text{MP}$ and ${\bf K}_\text{MP}$ token embeddings of so-called \leicr{coupled-mode} tokens. 

\vspace{0.1cm}
\noindent{\bf Weighted pooling.} Attention layer in Eq.~\eqref{eq:order-attn} produces feature representation ${\bf O}_\text{MP}\!\in\!\mbr{d''\times N}$ to enhance the relationship between for example feature channels and body joints. Subsequently, we handle the impact of hyper-edges of multiple orders by weighted pooling along hyper-edges of order $m\in\idx{r}$: 
\begin{align}
    & {\bf O}_\text{MP}^{*(m)}\!=\!{\bf O}_\text{MP}^{(m)}{\bf H}^{(m)}\!\in\! \mbr{d''\times J}, 
    \label{eq:ord-wise}
\end{align}
where ${\bf O}_\text{MP}^{(m)}\!\in\!\mbr{d''\times N_{E_m}}$ is simply extracted from ${\bf O}_\text{MP}$ for hyper-edges of order $m$, and  matrices ${\bf H}^{(m)}\!\in\!\mbr{N_{E_m}\times J}$ are learnable weights to perform weighted pooling along hyper-edges of order $m$. Finally, we obtain ${\bf O}_\text{MP}^{*}\!\in\!\mbr{r{d''\times J}}$ by simply concatenating ${\bf O}_\text{MP}^{*(1)},\cdots,{\bf O}_\text{MP}^{*(r)}$. If we used the input to MP from TP, then we denote the output of MP as ${{\bf O}'}_\text{MP}^{*}$.
%


\subsubsection{\lei{Temporal block Pooling (TP) Module}}

\vspace{0.1cm}
\noindent{\bf \lei{\leicr{CmSA}} in TP.} 
Firstly, we  reshape the multi-order feature representation $\lei{\tM}\!\in\!\mbr{d'\!\times\!N\!\times\!\tau}$ into ${\bf M}\!\in\!\mbr{d'N\!\times\!\tau}$ (or reshape the output from MP into ${\bf M}''\!\in\!\mbr{rd'J\!\times\!\tau}$).
For simplicity, we denote $d'''\!=\!d'N$ in the first case and $d'''\!=\!rd'J$ in the second case. As the first mode of reshaped input serves to form tokens, they are again \leicr{coupled-mode} tokens, \eg, `channel-hyper-edge' and `order-channel-\lei{body} joint' tokens, respectively.
Moreover, TP also performs pooling along block-temporal mode (along $\tau$). We form \lei{an \leicr{coupled-mode} self-attention}:
\begin{equation}
    a_\text{TP}({\bf Q}_\text{TP}, {\bf K}_\text{TP}, {\bf V}_\text{TP}) \!=\!\text{SoftMax}\left(\frac{{\bf Q}_\text{TP}{\bf K}_\text{TP}^\top}{\sqrt{d_{K_\text{TP}}}}\right)\!{\bf V}_\text{TP},
    \label{eq:temp-attn}
\end{equation}
where $\sqrt{d_{K_\text{TP}}}$ is the scaling factor, ${\bf Q}_\text{TP}\!=\!{\bf W}_\text{TP}^q{\bf M}$, ${\bf K}_\text{TP}\!=\!{\bf W}_\text{TP}^k{\bf M}$ and ${\bf V}_\text{TP}\!=\!{\bf W}_\text{TP}^v{\bf M}$ (we can use here ${\bf M}$ or ${\bf M}''$) are the query, key and value. Moreover, ${\bf Q}_\text{TP}$, ${\bf K}_\text{TP}$, ${\bf V}_\text{TP}\!\in\!\mbr{d'''\times \tau}$ 
and ${\bf W}_\text{TP}^q$, ${\bf W}_\text{TP}^k$, ${\bf W}_\text{TP}^v\!\in\!\mbr{d'''\times d'''}$ 
are learnable weights.  Eq.~\eqref{eq:temp-attn} reweighs  ${\bf V}_\text{TP}$ based on the correlation between ${\bf Q}_\text{TP}$ and ${\bf K}_\text{TP}$ token embeddings of \leicr{coupled-mode} tokens (`channel-hyper-edge' or `order-channel-body joint'). 
The output of attention is the temporal representation ${\bf O}_\text{TP} \!\in\! \mbr{d'''\times \tau}$. 
If we used ${\bf M}''$ as input, we denote the output as ${\bf O}''_\text{TP}$.

\vspace{0.1cm}
\noindent{\bf Pooling step.} Given the temporal representation ${\bf O}_\text{TP}\!\in\!\mbr{d'''\!\times\!\tau}$ (or ${\bf O}''_\text{TP}$), 
we apply pooling along the block-temporal mode to obtain compact feature representations independent of length (block count $\tau$) of skeleton sequence. There exist many pooling operations\footnote{We do not propose pooling operators but we  select popular ones with the purpose of comparing their impact on TP.} including first-order, \eg, average, maximum, sum pooling, second-order~\cite{Gao_2019_CVPR,NEURIPS2018_17c276c8} such as attentional pooling~\cite{Girdhar_17b_AttentionalPoolingAction}, higher-order (tri-linear)~\cite{7926605, koniusz2021high} and rank pooling~\cite{10.1109/TPAMI.2016.2558148}. The output  after pooling is ${\bf O}^*_\text{TP}\!\in\!\mbr{d'''}\!$ (or ${{\bf O}''}^*_\text{TP}$). 



\subsubsection{\lei{Model Variants}}

We devise four model variants by different stacking of MP with TP, with the goal of exploiting attention with different kinds of \leicr{coupled-mode} tokens:
\begin{enumerate}
    \item Single-branch: MP followed by TP, denoted MP$\rightarrow$TP, (Fig.~\ref{fig:pipeline} top right branch).
    \item Single-branch: TP followed by MP, denoted TP$\rightarrow$MP,  
    (Fig.~\ref{fig:pipeline} bottom right branch). 
    \item Two-branch (\lei{our 3Mformer,} Fig.~\ref{fig:pipeline}) which concatenates outputs of  MP$\rightarrow$TP and TP$\rightarrow$MP.
    \item We also investigate only MP or TP module followed by average pooling or an FC layer.
\end{enumerate}
The outputs from MP$\rightarrow$TP and TP$\rightarrow$MP have exactly the same feature dimension ($\mbr{rd'J}$, after reshaping into vector). For two-branch (\lei{our 3Mformer}), we simply concatenate these outputs ($\mbr{2rd'J}$, after concatenation). These vectors are forwarded to the FC layer to learn a classifier.
%

\section{Experiments}

\subsection{Datasets and Protocols} 

\noindent{{\bf (i) NTU RGB+D (NTU-60)}}~\cite{Shahroudy_2016_NTURGBD} contains 56,880 video sequences.
This dataset has variable sequence lengths  and high intra-class variations. Each skeleton sequence has 25 joints and there are no more than two human subjects in each video. Two evaluation protocols are: (i) cross-subject (X-Sub) and (ii) cross-view (X-View).

\noindent{{\bf (ii) NTU RGB+D 120 (NTU-120)}}~\cite{Liu_2019_NTURGBD120}, an extension of NTU-60, contains 120 action classes (daily/health-related), and 114,480 RGB+D video samples  captured with 106 distinct human subjects from 155 different camera viewpoints. There are also two evaluation protocols: (i) cross-subject (X-Sub) and (ii) cross-setup (X-Set).

\noindent{{\bf (iii) Kinetics-Skeleton}}, based on Kinetics~\cite{kay2017kinetics}, is  large-scale dataset with 300,000 video clips and  up to 400 human actions collected from YouTube. This dataset involves human daily activities, sports scenes and complex human-computer interaction scenes. Since Kinetics only provides raw videos without the skeletons, ST-GCN~\cite{stgcn2018aaai} uses the publicly available OpenPose toolbox~\cite{Cao_2017_CVPR} to estimate and extract the location of 18 human body joints on every frame in the clips. We use their released skeleton data to evaluate our model. Following the standard evaluation protocol, we report the Top-1 and Top-5 accuracies on the validation set.

\noindent{{\bf (iv) Northwestern-UCLA}}~\cite{wang2014cross} was captured by 3 Kinect cameras simultaneously from multiple viewpoints. It contains 1494 video clips covering 10 actions. Each action is performed by 10 different subjects. We follow the same evaluation protocol as~\cite{wang2014cross}: %
training split is formed from the first two cameras, and testing split from the last camera.


\subsection{Experimental Setup} 
We use PyTorch and 1$\times$Titan RTX 3090 for experiments. We use the Stochastic Gradient Descent (SGD)  with momentum 0.9, cross-entropy as the loss, weight decay of 0.0001 and batch size of 32. The learning rate is set to 0.1 initially. On NTU-60 and NTU-120, the learning rate is divided by 10 at the 40th and 50th epoch, and the training process ends at the 60th epoch. On Kinetics-Skeleton, the learning rate is divided by 10 at the 50th and 60th epoch, and the training finishes at the 80th epoch.
{We took 20\% of training set for validation to tune hyperparameters.}
All models have fixed hyperparameters with 2 and 4 layers for NTU-60/NTU-120 and Kinetics-Skeleton, respectively. The hidden dimensions is set to 16 for all 3 datasets. 
We use 4 attention heads for NTU-60 and NTU-120, and 8 attention heads for Kinetics-Skeleton.
To form each video temporal block, we simply choose temporal block size to be 10 and stride to be 5 to allow a 50\% overlap between consecutive temporal blocks. 
{For Northwestern-UCLA, the batch size is 16. We adopted the data pre-processing in~\cite{cheng2020shiftgcn}.}

\subsection{Ablation Study}
\noindent{\bf Search for the single best order $n$.} \lei{Table~\ref{tab:order} shows our analysis regarding the best order $n$. In general, increasing the order $n$ improves the performance (within $\sim$ 0.5\% on average), but causing higher computational cost, \eg, the number of hyper-edges for the skeletal hypergraph of order $n\!=\!4$ is 3060 on Kinetics-Skeleton. We also notice that combining orders 3 and 4 yields very limited improvements. The main reasons are: (i) reasonable order $n$, \eg, $n = 3$ or 4 improves accuracy as higher-order motion patterns are captured which are useful for classification-related tasks (ii) further increasing order $n$, \eg, $n = 5$ introduces patterns  in feature representations that rarely repeat even for the same action class.} Considering the cost and performance, we choose the maximum order $r\!=\!3$ ($n=1, 2, 3$) in the following experiments unless specified otherwise.

\begin{table}[t!]
\caption{Search for the single best order $n$ of hypergraph (except for $n\!=\!3\,\&\,4$ where we check if  $n\!=\!3\,\&\,4$ are complementary).}
\vspace{-0.5cm}
\begin{center}
\resizebox{0.9\linewidth}{!}{\begin{tabular}{l c c c c c}
\toprule
\multirow{2}{*}{Order-$n$}& \multicolumn{2}{c}{NTU-60} & \multicolumn{2}{c}{NTU-120} & Kinetics-Skel.\\
\cline{2-6}
& {X-Sub} & {X-View} & {X-Sub} & {X-Set} & Top-1 acc.\\
\midrule
$n=1$ & 78.5 & 86.3 & 75.3 & 77.9 & 32.0\\
$n=2$ & 83.0 & 89.2 & 86.2 & 88.3 & 37.1\\
$n=3$ & 91.3 & 97.0 & 87.5 & 89.7 & 39.5\\
$n=4$ & 91.5 & 97.1 & {\bf 87.8} & 90.0 & 40.1\\
$n=5$ & 91.4 & {\bf 97.3} & {\bf 87.8} & 90.0 & 40.3\\
$n=3~\&~4$& {\bf 91.6} & 97.2 & 87.6 & {\bf 90.3} & {\bf 40.5}\\
\bottomrule
\end{tabular}}
\label{tab:order}
\end{center}
\end{table}

\begin{table}[t!]
\caption{Evaluations of our model variants with/without MP and/or TP. \lei{Baseline in the table denotes the backbone (MLP unit + HoTs) without the use of either MP or TP module.}}
\vspace{-0.5cm}
\begin{center}
\resizebox{\linewidth}{!}{\begin{tabular}{l c c c c c}
\toprule
\multirow{2}{*}{Variants}& \multicolumn{2}{c}{NTU-60} & \multicolumn{2}{c}{NTU-120} & Kinetics-Skel.\\
\cline{2-6}
& {X-Sub} & {X-View} & {X-Sub} & {X-Set} & Top-1 acc.\\
\midrule
\lei{\quad Baseline} & 89.8 & 91.4 & 86.5 & 87.0 & 38.6\\
+ TP only& 91.2 & 93.8 & 87.5 & 88.6 & 39.8\\
+ MP only& 92.0 & 94.3 & 88.7 & 89.7 & 40.3 \\
+ MP$\rightarrow$TP & 93.0 & 96.1 & 90.8 & 91.7 & 45.7 \\
+ TP$\rightarrow$MP & 92.6 & 95.8 & 90.2 & 91.1 & 44.0 \\
\rowcolor{blue!10}
+ 2-branch(3Mformer)$\!\!\!\!$ & {\bf 94.8} & {\bf 98.7} & {\bf 92.0} & {\bf 93.8} & {\bf 48.3}\\
\bottomrule
\end{tabular}}
\label{tab:cross-joint-attn}
\end{center}
\end{table}


\begin{table*}[t!]
\caption{Experimental results on NTU-60, NTU-120 and Kinetics-Skeleton.}
\vspace{-0.5cm}
\begin{center}
\resizebox{\linewidth}{!}{\begin{tabular}{l l l c c c c c c c c}
\toprule
 & \multirow{2}{*}{Method} & \multirow{2}{*}{Venue}  & \multicolumn{2}{c}{NTU-60} & & \multicolumn{2}{c}{NTU-120} & & \multicolumn{2}{c}{Kinetics-Skeleton}\\
\cline{4-5}
\cline{7-8}
\cline{10-11}
& & & {X-Sub} & {X-View} && {X-Sub} & {X-Set}  && {Top-1} & {Top-5} \\
\midrule
\multirow{10}{*}{\parbox{2.0cm}{\bf Graph-based}} 
& TCN~\cite{8014941} & CVPRW'17 & - & - & & - & - && 20.3 & 40.0\\
& ST-GCN~\cite{stgcn2018aaai} & AAAI'18 & 81.5 & 88.3 && 70.7 & 73.2 &&30.7 & 52.8 \\
& AS-GCN~\cite{Li_2019_CVPR}  & CVPR'19 & 86.8 & 94.2 && 78.3 & 79.8 && 34.8 & 56.5\\
& 2S-AGCN~\cite{2sagcn2019cvpr} & CVPR'19 & 88.5 & 95.1&& 82.5 & 84.2 && 36.1 & 58.7\\
& NAS-GCN~\cite{Peng_Hong_Chen_Zhao_2020} & AAAI'20 & 89.4 & 95.7&& -& -&& 37.1 & 60.1\\
& Sym-GNN~\cite{9334430} & TPAMI'22 & 90.1 & 96.4&& - & - && 37.2 & 58.1\\
& Shift-GCN~\cite{cheng2020shiftgcn} & CVPR'20 & 90.7 & 96.5 && 85.9 & 87.6&& - &-\\
& MS-G3D~\cite{Liu_2020_CVPR} & CVPR'20 & 91.5 & 96.2 & & 86.9 & 88.4 & & 38.0 & 60.9\\
& CTR-GCN~\cite{chen2021channel} & ICCV'21 & 92.4 & 96.8 & & 88.9 & 90.6 & & - & -\\
& InfoGCN~\cite{Chi_2022_CVPR} & CVPR'22 & 93.0 & 97.1 & & 89.8 & 91.2 & & - & -\\
& PoseConv3D~\cite{Duan_2022_CVPR} & CVPR'22 & 94.1 & 97.1 & & 86.9 & 90.3 & & {\bf 47.7} & - \\
\hline
\multirow{4}{*}{\parbox{2.0cm}{\bf Hypergraph-based}}
& Hyper-GNN~\cite{9329123} & TIP'21 & 89.5 & 95.7&& - &-  && 37.1 & 60.0\\
& DHGCN~\cite{dynamichypergraph} & CoRR'21 & 90.7 & 96.0&& 86.0 & 87.9 && 37.7 & 60.6\\
& Selective-HCN~\cite{10.1145/3512527.3531367}& ICMR'22 & 90.8 & 96.6 && - &- && 38.0 & 61.1\\
& SD-HGCN~\cite{10.1007/978-3-030-92270-2_2} & ICONIP'21 &  90.9 & 96.7&& 87.0 & 88.2 && 37.4 & 60.5\\
\hline
\multirow{9}{*}{\parbox{2.0cm}{\bf Transformer-based}}
& ST-TR~\cite{PLIZZARI2021103219}& CVIU'21 & 90.3 & 96.3&& 85.1 & 87.1 && 38.0 &60.5\\
& MTT~\cite{9681250}& LSP'21 & 90.8 & 96.7 && 86.1 & 87.6 && 37.9 &61.3\\
& 4s-GSTN~\cite{sym14081547} & Symmetry'22 & 91.3 & 96.6 && 86.4 & 88.7 && -&-\\
& STST~\cite{10.1145/3474085.3475473} &  ACM MM'21& 91.9 & 96.8 && -&- && 38.3 &61.2 \\
\cdashline{2-11}
& 3Mformer (with avg-pool, {\em ours}) & &  92.0 & 97.3&& 88.0 & 90.1 && 43.1&65.2\\
& 3Mformer (with max-pool, {\em ours}) &  & 92.1 & 97.8&& - & - && -&-\\
\rowcolor{blue!10}
& 3Mformer (with attn-pool, {\em ours}) & & {\bf 94.2} & {\bf 98.5}&& 89.7 & 92.4 && 45.7 & 67.6\\
\rowcolor{blue!10}
& 3Mformer (with tri-pool, {\em ours}) & & 94.0 & {\bf 98.5}&& {\bf 91.2} & {\bf 92.7} && {\bf 47.7} & {\bf 71.9}\\
\rowcolor{blue!10}
& 3Mformer (with rank-pool, {\em ours}) & & {\bf 94.8} & {\bf 98.7}&& {\bf 92.0} & {\bf 93.8} && {\bf 48.3} & {\bf 72.3}\\
\bottomrule
\end{tabular}}
\label{tab:globaltable}
\end{center}
\end{table*}

\begin{figure}[t]
\centering
\begin{subfigure}[t]{0.22\linewidth}
\centering\includegraphics[width=\linewidth,height=1.9cm]{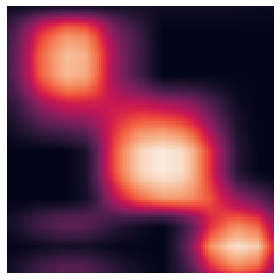}
\caption{}\label{fig:temp}
\end{subfigure}\hfill
\begin{subfigure}[t]{0.22\linewidth}
\centering\includegraphics[width=\linewidth,height=1.9cm]{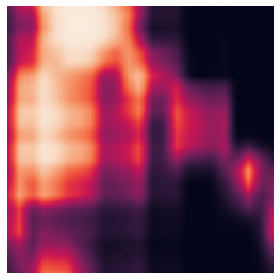}
\caption{}\label{fig:channel-hyper-edge}
\end{subfigure}\hfill
\begin{subfigure}[t]{0.22\linewidth}
\centering\includegraphics[width=\linewidth,height=1.9cm]{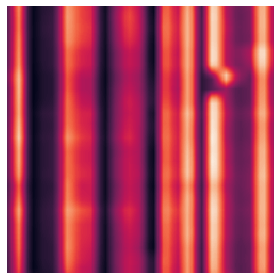}
\caption{}\label{fig:order-channel}
\end{subfigure}\hfill
\begin{subfigure}[t]{0.305\linewidth}
\centering\includegraphics[width=\linewidth,height=1.9cm]{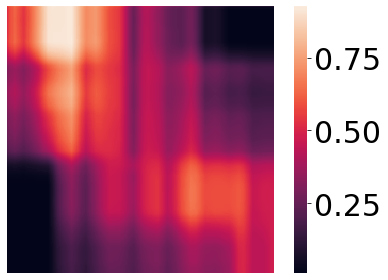}
\caption{}\label{fig:channel-temp}
\end{subfigure}
\caption{\lei{Visualization of attention matrices. (a) single-mode attention matrix of `channel-only' token, (b)--(d) \leicr{coupled-mode} attention matrices of `channel-hyper-edge', `order-channel-body joint' and `channel-temporal block' tokens, respectively.}}
\label{fig:attn-matrix}
\end{figure}

\begin{figure}[t]
\centering\includegraphics[width=1\linewidth]{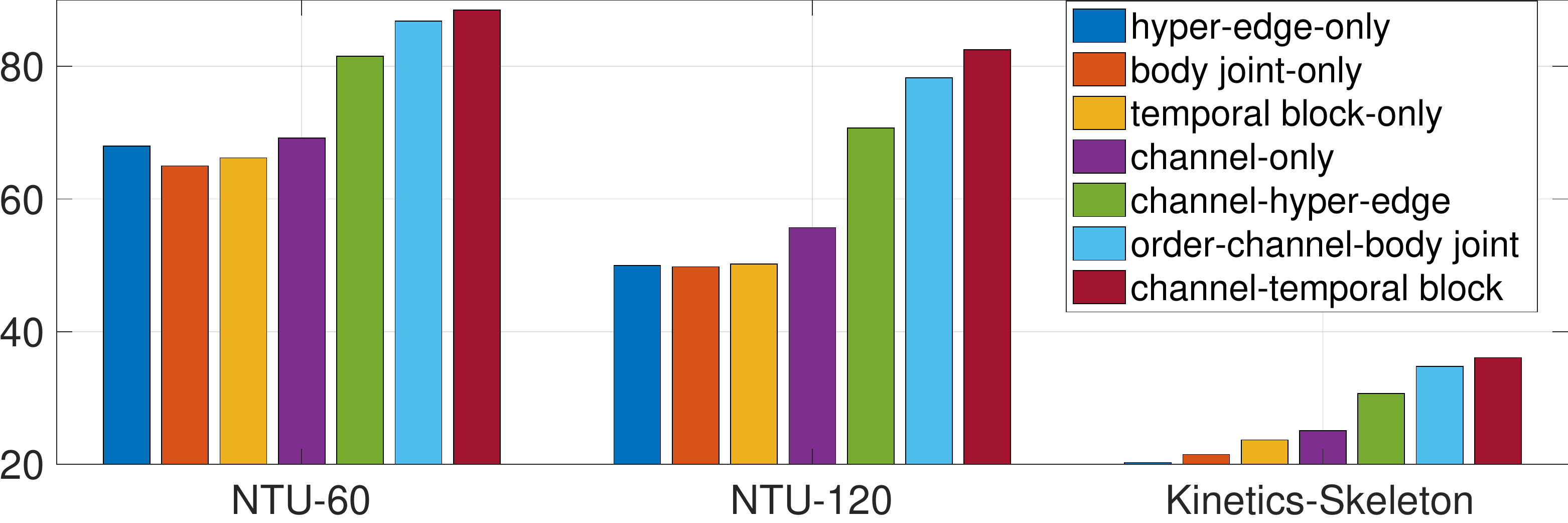}
\caption{\lei{Evaluations of different single-mode ({\it baseline}) and \leicr{coupled-mode} tokens. We use a 3rd-order HoT with a standard Transformer, but we replace the scaled dot-product attention with \leicr{coupled-mode} tokens and \leicr{coupled-mode} attention.}}
\label{fig:token-comp}
\end{figure}

\vspace{0.1cm}
\noindent\lei{{\bf Discussion on \leicr{coupled-mode} attention.} Fig.~\ref{fig:attn-matrix} shows the visualization of some attention matrices in our 3Mformer, which show diagonal and/or vertical patterns that are consistent with the patterns of the attention matrices found in standard Transformer trained on sequences,  \eg, for natural language processing tasks~\cite{NIPS2017_3f5ee243, kovaleva-etal-2019-revealing}. We also notice that the \leicr{coupled-mode} attention, \eg, `channel-temporal block' captures much richer information compared to single mode attention, \eg, `channel-only'. Our \leicr{coupled-mode} attention can be applied to different orders of tensor representations through simple matricization.}

\vspace{0.1cm}
\noindent{\bf \lei{Discussion on model variants}.} To show the effectiveness of the proposed MP and TP module, firstly, we compare TP only and MP only with the baseline (\lei{No MP or TP module}). We use the TP module followed by an FC layer instead of MP as in TP$\rightarrow$MP, where the FC layer takes the output from TP ($\mbr{d'N}$) and produces a vector in $\mbr{3d'J}$, passed to the classifier. Similarly, for MP only, we use the MP module followed by an average pooling layer instead of TP as in MP$\rightarrow$TP, where the average layer takes output from MP ($\mbr{3d'J\times \tau}$) and generates a vector in $\mbr{3d'J}$ (pool along $\tau$ blocks), passed to the classifier.
Table~\ref{tab:cross-joint-attn} shows the results. 
\lei{With just the TP module, we outperform the baseline by 1.3\% on average. With only the MP module, we outperform the baseline by 2.34\% on average. These comparisons show that (i) \leicr{CmSA} in MP and TP are efficient for better performance (ii) MP performs better than TP which shows that `channel-temporal block' token contains richer information than `channel-hyper-edge' token. 
We also notice that MP$\rightarrow$TP slightly outperforms TP$\rightarrow$MP by $\sim$ 1\%, and the main reason is that MP$\rightarrow$TP has \leicr{coupled-mode} tokens `channel-temporal block' and `order-channel-joint' which attend 4 modes, whereas TP$\rightarrow$MP has `channel-hyper-edge' and `channel-only' tokens which attend only 2 modes. Fig.~\ref{fig:token-comp} shows a comparison of different \leicr{coupled-mode} tokens on 3 benchmark datasets. This also suggests that one should firstly perform attention with \leicr{coupled-mode} `channel-block' tokens, followed by weighted pooling along the hyper-edge mode, followed by attention with \leicr{coupled-mode} `order-channel-body joint' and finalised by block-temporal pooling. Finally, with 2-branch (3Mformer), we further boost the performance by 2--4\%, which shows that MP$\rightarrow$TP and TP$\rightarrow$MP are complementary branches. Below we use 2-branch (3Mformer) in the experiments (as in Fig.~\ref{fig:pipeline}). 
}
%




\vspace{0.1cm}
\noindent{\bf Comparison of pooling in TP.} As shown in Table~\ref{tab:globaltable}, average pooling (avg-pool) achieves similar performance (within $\sim$ 0.5\% difference) as maximum pooling (max-pool), second-order pooling (attn-pool) outperforms average and maximum pooling by $\sim$ 1--2\% and third-order pooling (tri-pool) outperforms second-order pooling by $\sim$ 1\%. Interestingly, rank pooling (rank-pool) achieves the best performance. We think it is reasonable as rank pooling strives to enforce the temporal order in the feature space to be preserved, \eg, it forces network to always preserve temporal progression of actions over time. With multiple attention modules, orderless statistics such as second- or third-order pooling may be too general.

\subsection{Comparisons with the State of the Arts}

We compare our model with recent state-of-the-art methods. On the NTU-60 (Tab.~\ref{tab:globaltable}), we obtain the top-1 accuracies of the two evaluation protocols during test stage. The methods in comparisons include popular graph-based~\cite{stgcn2018aaai,Li_2019_CVPR,2sagcn2019cvpr,Peng_Hong_Chen_Zhao_2020,9334430} and hypergraph-based models~\cite{9329123,dynamichypergraph,10.1145/3512527.3531367,10.1007/978-3-030-92270-2_2}. Our 3rd-order model  outperforms all graph-based methods, and also outperforms existing hypergraph-based models such as Selective-HCN and SD-HGCN by 0.45\% and 0.35\% on average on X-Sub and X-View respectively. With \lei{3Mformer} for the fusion of multi-order features, our model further boosts the performance by $\sim$ 3\% and 1.5\% on the two protocols.

It can be seen from Tab.~\ref{tab:globaltable} on NTU-60 that although some learned graph-based methods such as AS-GCN and 2S-AGCN can also capture the dependencies between human body joints, they only consider the pairwise relationship between body joints, which is the second-order interaction, and ignore the higher-order interaction between multiple body joints in form of hyper-edges, which may lose sensitivity to important groups of body joints. Our proposed \lei{3Mformer} achieves better performance by constructing a hypergraph from 2D/3D body joints as nodes for action recognition, thus capturing  higher-order interactions of body joints to further improve the performance. Note that even with the average pooling, our model still achieves competitive results compared to its counterparts.

For the NTU-120 dataset (Tab.~\ref{tab:globaltable}), we obtain the top-1 performance on X-Sub and X-Set protocols. Our 2nd-order HoT alone outperforms graph-based models by 2--2.4\% on average. For example, we outperform recent Shift-GCN by 0.3\% and 0.7\% on X-Sub and X-Set respectively. Moreover, our 3rd-order HoT alone outperforms SD-HGCN by 0.5\% and 1.5\% respectively on X-Sub and X-Set. With the \lei{3Mformer} for the fusion of multi-order feature maps, we obtain the new state-of-the-art results. \leicr{Notice that our 3Mformer yields 92.0\% / 93.8\% on NTU-120 while~\cite{peng2021rethinking} yields 80.5\% / 81.7\% as we explore the fusion of multiple orders of hyperedges and several coupled-token types capturing easy-to-complex dynamics of varying joint groups.}

As videos from the Kinetics dataset are processed by the OpenPose, the skeletons in the Kinetics-Skeleton dataset have defects which adversely affect the performance of the model. We show both top-1 and top-5 performance in Table~\ref{tab:globaltable} to better reflect the performance of our \lei{3Mformer}. ST-GCN is the first method based on GCN, our 2nd-order HoT alone achieves very competitive results compared to the very recent NAS-GCN and Sym-GNN. The 3rd-order HoT alone outperforms Hyper-GNN, SD-HGCN and Selective-HCN by 3.4\%, 3.1\% and 2.9\% respectively for top-1 accuracies. Moreover, fusing multi-order feature maps from multiple orders of hyper-edges via \lei{3Mformer} gives us the best performance on Kinetics-Skeleton with 48.3\% for top-1, the new state-of-the-art result.


\begin{table}[t!]
\caption{Experimental results on Northwestern-UCLA.}
\vspace{-0.5cm}
\begin{center}
\resizebox{\linewidth}{!}{\begin{tabular}{l c c c c c c}
\toprule
& Shift-GCN~\cite{cheng2020shiftgcn} & CTR-GCN~\cite{chen2021channel} & InfoGCN~\cite{Chi_2022_CVPR} & 2nd-order & 3rd-order& 3Mformer\\
& (CVPR'20) & (ICCV'21) & (CVPR'22) & only (\em{ours})& only (\em{ours}) & (\em{ours})\\
\midrule
acc.(\%) & 94.6 & 96.5 & 97.0 & 96.5 & {\bf 97.2} & {\bf 97.8}\\
\bottomrule
\end{tabular}}
\label{northwestern-ucla}
\end{center}
\end{table}
 
Table~\ref{northwestern-ucla} shows results on the Northwestern-UCLA dataset. Our 3Mformer is also effective on this dataset--it outperforms the current state-of-the-art InfoGCN by 0.8\%.

\section{Conclusions}

In this paper, we model the skeleton data as hypergraph to capture higher-order information formed between groups of human body joints of orders $1,\cdots,r$. We use Higher-order Transformer (HoT) to learn higher-order information on hypergraphs of  $r$-order formed over 2D/3D human body joints. We also introduce a novel \lei{Multi-order Multi-mode Transformer (3Mformer)} for the fusion of multi-order  feature representations. Our end-to-end trainable \lei{3Mformer} outperforms state-of-the-art graph- and hypergraph-based models by a large margin on several benchmarks. 

\paragraph{Acknowledgements.} 
LW is supported by the Data61/ CSIRO PhD Scholarship. 
PK is in part funded by CSIRO's Machine Learning and Artificial Intelligence Future Science Platform (MLAI FSP) Spatiotemporal Activity.

%

{\small
\bibliographystyle{ieee_fullname}
\bibliography{egbib}
}

\clearpage
\appendix


\title{\lei{3Mformer: } Multi-order Multi-mode Transformer \\for Skeletal Action Recognition \\-- Supplementary Material --}


\author{Lei Wang\textsuperscript{$\dagger,\S$}\qquad Piotr Koniusz\textsuperscript{$*, \S,\dagger$}\\
$^{\dagger}$Australian National University, $^\S$Data61\heart CSIRO\\
$^\S$firstname.lastname@data61.csiro.au
}

\maketitle



\section{Visualization of 3Mformer}
\label{app:3mf}

\begin{figure*}[ht]
  \centering
  \begin{subfigure}{\linewidth}
    \centering
    \includegraphics[trim=1.8cm 2.5cm 1.8cm 3.0cm, clip=true, width=\linewidth]{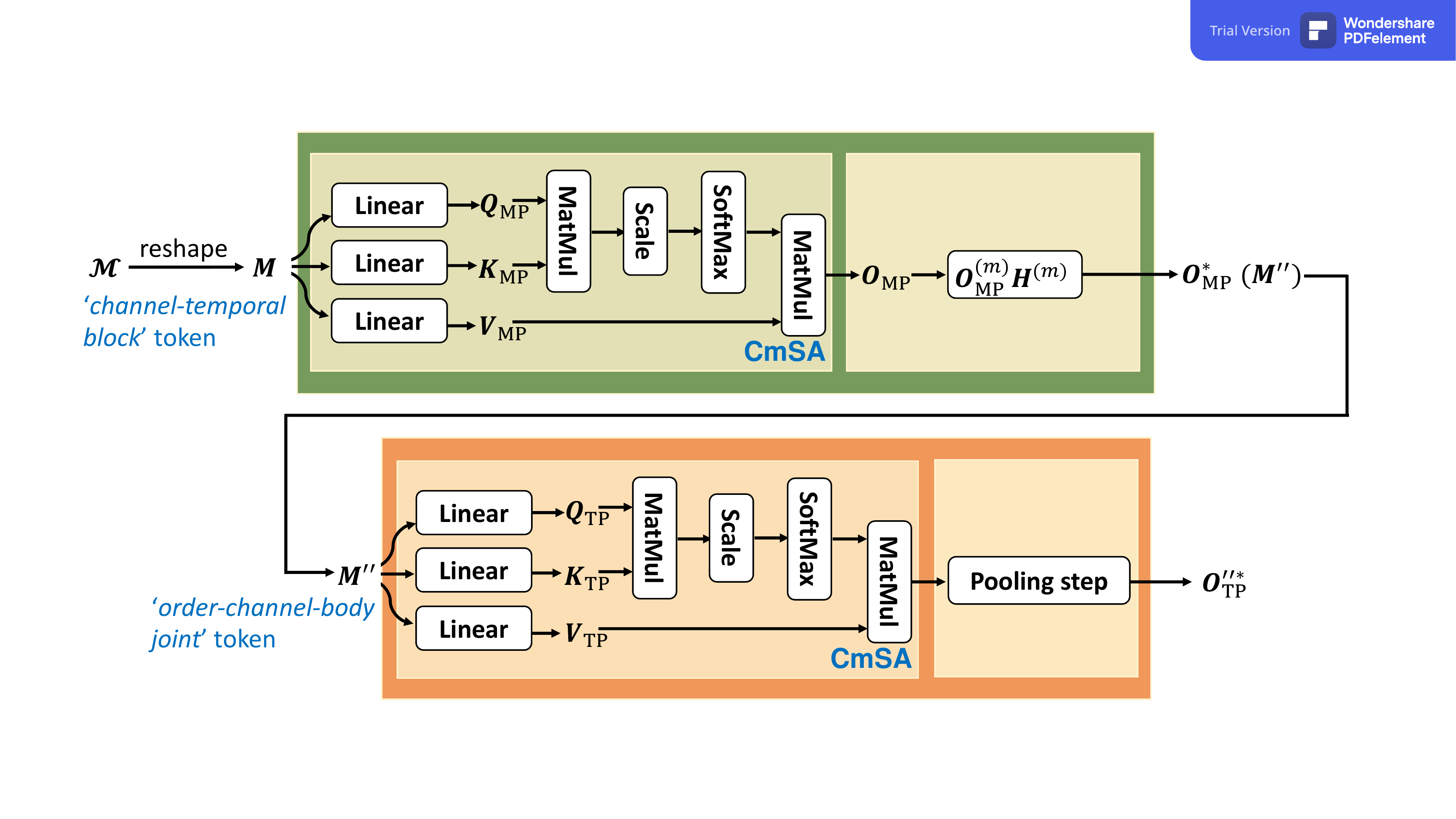}
    \caption{Single-branch: MP followed by TP (denoted MP$\rightarrow$TP).}
    \label{fig:mp-tp}
  \end{subfigure}
  \begin{subfigure}{\linewidth}
    \centering
    \includegraphics[trim=1.5cm 2.5cm 1.8cm 3.3cm, clip=true, width=\linewidth]{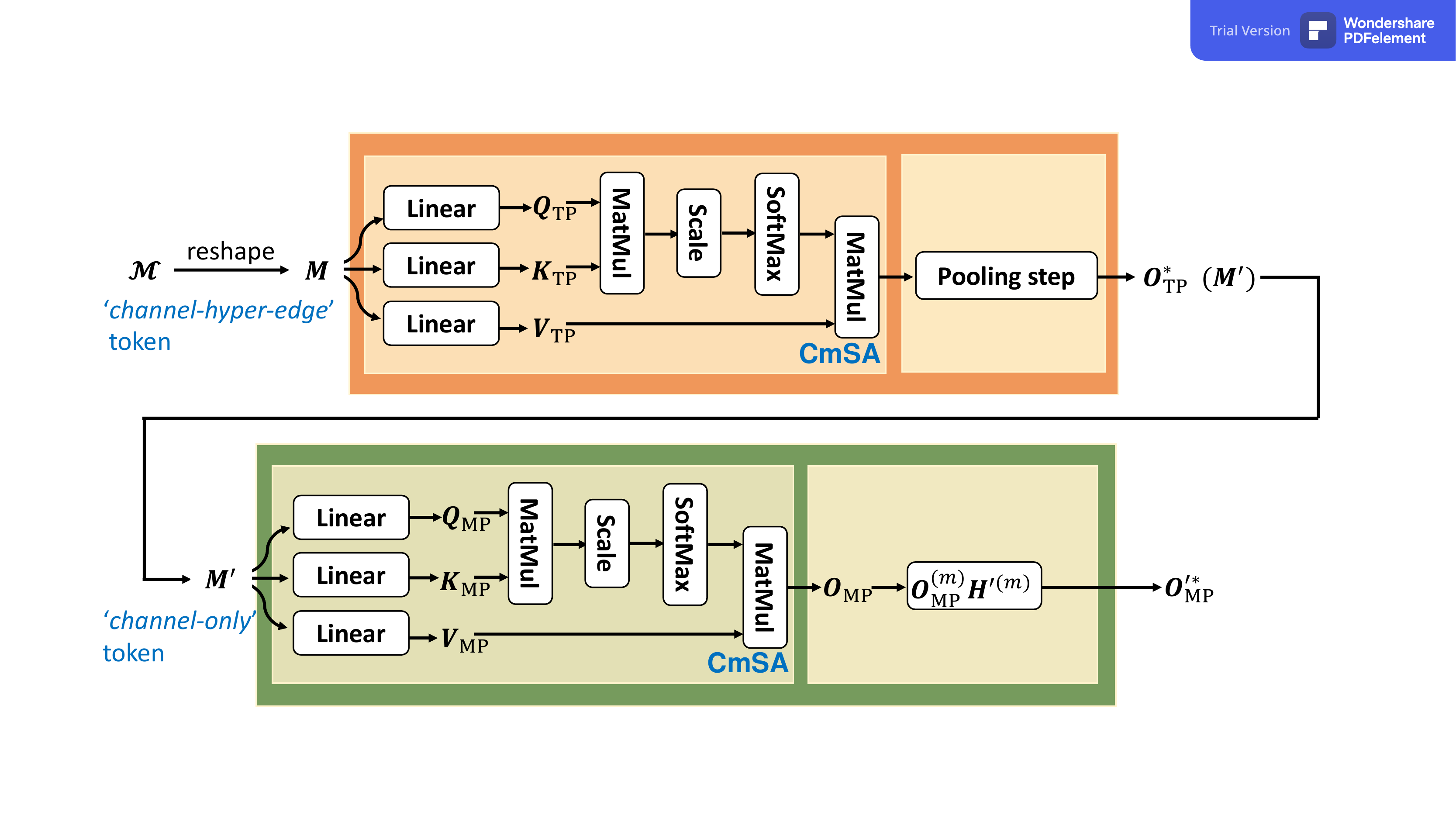}
    \caption{Single-branch: TP followed by MP (denoted TP$\rightarrow$MP).}
    \label{fig:tp-mp}
  \end{subfigure}  
  \caption{Visualization of 3Mformer which is a two- branch model: (a) MP$\rightarrow$TP and (b) TP$\rightarrow$MP. Green and orange blocks are Multi-order Pooling (MP) module and Temporal block Pooling (TP) module, respectively. $(m)$ inside the MP module denotes the order $m\in\idx{r}$ of hyper-edges. These two modules (MP and TP) are the basic building blocks which are further stacked to form our 3Mformer. Each module (MP or TP) uses a specific \leicr{coupled-mode} token through matricization (we use reshape for simplicity), \eg, `channel-temporal block', `order-channel-body joint', `channel-hyper-edge (any order)' or `channel-only', and the \leicr{Coupled-mode} Self-Attention (\leicr{CmSA}) is used to explore the \leicr{coupled-mode} relationships inside the \leicr{coupled-mode} tokens. We also form our multi-head \leicr{CmSA} as in standard Transformer (where the \leicr{CmSA} module repeats its computations multiple times in parallel and the attention module splits the query, key and value, each split is independently passed through a separate head and later combined together to produce the final \leicr{coupled-mode} attention score). We omit the multi-head visualization for simplicity and better visualization purposes. }  
  \label{fig:3mformer}
\end{figure*}

Fig.~\ref{fig:3mformer} shows the visualization of our 3Mformer. The green and orange blocks denote the Multi-order Pooling (MP) and the Temporal block Pooling (TP) respectively, which are two basic building blocks that can be stacked to form our 3Mformer. More precisely, our 3Mformer consists of two branches: (i) MP followed by TP (denoted MP $\rightarrow$ TP, Fig.~\ref{fig:mp-tp}) and (ii) TP followed by MP (denoted TP $\rightarrow$ MP, Fig.~\ref{fig:tp-mp}). 

\section{Skeletal Graph and Hypergraph}

\noindent\textbf{Skeletal Graph}~\cite{stgcn2018aaai}. Let $G\!=\!({V}, {E})$ be a skeletal graph with the vertex set ${V}$  of nodes (body joints) $\{v_1, \cdots, v_J\}$, and ${E}$ be edges (bones) of the graph, and ${E}$ consists of ${ E}_S$ and ${E}_T$. The subset ${E}_S\!=\!\{(v_{it}, v_{jt}): i,j\!\in\!\idx{J} \text{ and } t\!\in\!\idx{T}\}$ represents that at time step $t$, each pair of joints $(v_{it}, v_{jt})$ corresponding to skeletal connectivity diagram is connected; whereas ${E}_T\!=\!\{(v_{it}, v_{i(t+1)}): i\!\in\!\idx{J} \text{ and } t\!\in\!\idx{T}\}$ forms the connection of the same joint across time. The set of joints and edges together form the skeleton graph. 
If two body joints are connected by an edge, the corresponding  element in the incidence matrix ${\bf H}$ is equal to 1, otherwise it is equal to 0, and the adjacency matrix ${\bf A}={\bf H}^T{\bf H}-2\mathbf{I}$ (where {\bf I} is the identity matrix). The update rule of a common GCN model at time step $t$ is defined as:
\begin{equation}
    {\bf X}_t^{(l+1)}\!=\!\sigma\big(\widetilde{\bf D}^{-\frac{1}{2}}\widetilde{\bf A}\widetilde{\bf D}^{-\frac{1}{2}}{\bf X}_t^{(l)}\mP^{(l)}\big),
\end{equation}
where $\sigma(\cdot)$ is a non-linearity, $\widetilde{\bf D}$ is the graph degree matrix, ${\bf X}_t^{(l)}$ is the input data of the convolutional layer $l$ at the time step $t$ and $\mP^{(l)}$ is the learnable parameters of layer $l$. $\widetilde{\bf A}\!=\!{\bf A}+{\bf I}$ is a normalized graph adjacency matrix.

The tensor representation of graph data can be given by ${\bf X} \!\in\!\mbr{J^2\!\times\!d}$ where $d$ is the feature channel dimension.

\vspace{0.1cm}
\noindent\textbf{Skeletal Hypergraph}~\cite{ijcai2020-109,9329123}. Hypergraph captures  complex higher-order relationships by hyper-edges that connect more than two nodes (body joints). Each hyper-edge is a subset of all nodes. Let $G_h\!=\!({ V}_h, {E}_h, {W}_h)$ where ${V}_h$, ${E}_h$ and ${W}_h$ denote respectively the set of body joints, hyper-edges and the weights of hyper-edges.
Given $v \!\in\!{V}_h$ and $e\!\in\!{E}_h$, the elements in the incidence matrix ${\bf H}_h$ of the skeleton hypergraph are defined as ${H}_{h,v,e}=1$, or simply put $h(v,e)=1$, if vertex $v$ is part of edge $e$, 0 otherwise.
The degree of node/body joint $v\!\in\!{V}_h$ is the number of hyper-edges passing through the node, which is defined as:
\begin{equation}
    d(v)\!=\!\sum_{e\in {E}_h}w(e)h(v, e),
\end{equation}
where $w(e)$ is the weight of hyper-edge $e$. The degree of hyper-edge $e\in {E}_h$ is the number of nodes (body joints) contained in the hyper-edge $e$ that satisfies:
\begin{equation}
    \delta(e)\!=\!\sum_{v\in {V}_h}h(v, e).
\end{equation}
Moreover, let ${\bf D}_v$ and ${\bf D}_e$ be the diagonal matrices of node degrees $d(v)$ and  the hyper-edge degrees $\delta(e)$ respectively. Let ${\bf W}$ denote the diagonal matrix of the hyper-edge weights (initially the weights of all hyper-edges are set to 1). Then the update rule of the Hypergraph Convolutional Network at the time step $t$ is given by:
\begin{equation}
    {\bf X}_t^{(l+1)}\!=\!\sigma\big({\bf D}_v^{\frac{1}{2}}{\bf H}_h{\bf W}{\bf D}_e^{-1}{\bf H}_h^\top{\bf D}_v^{\frac{1}{2}}{\bf X}_t^{(l)}\mP^{(l)}\big),
\end{equation}
where $\mP^{(l)}$ are learnable parameters for layer $l$.

\section{\lei{Skeleton Data Preprocessing}}

\lei{Before passing the skeleton sequences into MLP, we first normalize each body joint \wrt to the torso joint ${\bf v}_{f, c}$:
\begin{equation}
    {\bf v}^\prime_{f, i}\!=\!{\bf v}_{f, i}\!-\!{\bf v}_{f, c},
\end{equation}
where $f$ and $i$ are the index of video frame and human body joint respectively. After that, we further normalize each joint coordinate into  [-1, 1] range:
\begin{equation}
    \hat{{\bf v}}_{f, i}[j] = \frac{{\bf v}^\prime_{f, i}[j]}{ \text{max}([\text{abs}({\bf v}^\prime_{f, i}[j])]_{f\in{\mathcal{I}_\tau},i\in\mathcal{I}_{J} } )},
\end{equation}
where $j$ is for selection of the $x$, $y$ and $z$ axes, $\tau$ is the number of frames and $J$ is the number of 3D body joints per frame.}

\lei{For the skeleton sequences that have more than one performing subject, (i) we normalize each skeleton separately, and each skeleton is passed to MLP for learning the temporal dynamics, and (ii) for the output features per skeleton from MLP, we pass them separately to the block-temporal HoT, \eg, two skeletons from a given video sequence will have two outputs obtained from the the block-temporal HoT, and we aggregate the outputs through average pooling before passing our 3Mformer. }

\section{\leicr{Additional Results and Discussions}}

\subsection{\leicr{Ablations of MP}}

\leicr{We choose average pooling (avg-pool) and max-pooling (max-pool) for hyper-edge features in comparison to our learned weighted pooling (wei-pool), and the comparisons are given in Table~\ref{tab:wp-mp}. As shown in the table, our learned weighted pooling (wei-pool) consistently achieves the best performance on all 3 datasets.}
\begin{table}[t!]
\caption{Ablations of different pooling methods in MP.}
\vspace{-0.5cm}
\begin{center}
\resizebox{0.9\linewidth}{!}{\begin{tabular}{l c c c c c}
\toprule
\multirow{2}{*}{Pooling}& \multicolumn{2}{c}{NTU-60} & \multicolumn{2}{c}{NTU-120} & Kinetics-Skel.\\
\cline{2-6}
& {X-Sub} & {X-View} & {X-Sub} & {X-Set} & Top-1 acc.\\
\midrule
avg-pool & 91.3 & 96.8 & 86.5 & 89.0 & 41.9\\
max-pool & 92.7 & 98.0 & 88.5 & 91.0 & 43.8\\
\rowcolor{blue!10} wei-pool ({\it ours}) & 94.8 & 98.7 & 92.0 & 93.8 & 48.3\\
\bottomrule
\vspace{-1.0cm}
\end{tabular}}
\label{tab:wp-mp}
\end{center}
\end{table}

\subsection{\leicr{Learning the short-term temporal patterns}}

\leicr{A block of $T$ neighbor frames are passed to the MLP unit to capture the short-term temporal patterns. The whole sequence consists of $\tau$ such blocks, each passed separately through the MLP unit (and each joint $1,\cdots,J$). Thus, the MLP only mixes the information from $1,\cdots,T$ frames of a given block/body joint $j$ and captures short-term relations (within-block) of a given 3D body joint (in contrast to between-block relations). The MLP unit input size is  $3T$; 3 due to 3D coordinate). The $\text{MLP}\!:\!\mbr{3T}\!\rightarrow\!\mbr{d}$ contains: FC ($3T\!\rightarrow\!6T$), ReLU, FC ($6T\!\rightarrow\!9T$), ReLU, Dropout, 
FC ($9T\!\rightarrow\!d$). $J$ body joints and $\tau$ blocks are treated as the batch dimension. Feature output size $d$: 100, 150, 420 on NTU-60, NTU-120, Kinetics-Skeleton.}

\subsection{\leicr{Why 3Mformer works and when does it fail?}}

\leicr{Our method works well as it (i) uses skeletal hypergraphs of various orders to learn the interaction of varying size groups of skeletal joints (as opposed to typical skeleton graph physical connectivity), (ii) fuses groups multiple orders by 3Mformer by several \leicr{coupled-token} types via two basic building blocks (MP \& TP) that learn various aspects of higher-order motion dynamics. Multiple-order hyperedges are more resistant to noise (\eg,  Kinetics-Skeleton is noisy due to the pose estimation errors), if one body joint is noisy (but the rest is stable). We inject Gaussian noise into 3D {\em ankle} joints, vary noise amplitude, and we show the experimental results in Table~\ref{tab:3mformer-gaussian}. As shown in the table, our 3Mformer copes with noise better than ST-GCN.}

\begin{table}[t!]
\caption{Comparisons of robustness \wrt Gaussian noise.}
\vspace{-0.5cm}
\begin{center}
\resizebox{0.85\linewidth}{!}{\begin{tabular}{l c c c c}
\toprule
 & original & $\times$ 1 & $\times$ 1.5 & $\times$ 2 \\
\midrule
ST-GCN & 81.5 &  74.9 (\textcolor{blue}{$\downarrow$6.6}) &  69.2 (\textcolor{blue}{$\downarrow$12.3}) & 50.1 (\textcolor{blue}{$\downarrow$31.4}) \\
3Mformer & 94.8 & 91.9 (\textcolor{blue}{$\downarrow$2.9}) & 89.5 (\textcolor{blue}{$\downarrow$5.3}) & 86.8 (\textcolor{blue}{$\downarrow$8.0})\\
\bottomrule
\vspace{-1.0cm}
\end{tabular}}
\label{tab:3mformer-gaussian}
\end{center}
\end{table}

\leicr{Our method may underperform if (i) the backbone encoder cannot efficiently produce higher-order features (ii) the number of skeletal joints are very large (the number of hyper-edge features would be very large) (iii) when dataset is too small to learn high-order interactions (extra learnable parameters). For example, see the experimental results on MSRAction3D in Table~\ref{3mformer-msr}.}

\leicr{We notice that small datasets may be not enough to train high-order models (Table~\ref{3mformer-msr}). On key classic large datasets, NTU-60, NTU-120 and Kinetics-Skeleton, we do not observe any issue as human motions exhibit similar multi-joint dynamics for typical action classes. Perhaps some fine-grained unusual action classes could pose problems.} 

\begin{table}[t!]
\caption{Experimental results on MSRAction3D.}
\vspace{-0.5cm}
\begin{center}
\resizebox{0.42\linewidth}{!}{\begin{tabular}{l c c c}
\toprule
order &  2 & 3 & 4\\
\midrule
acc.(\%) & 73.82 & 63.64 & 55.27\\
\bottomrule
\vspace{-1.0cm}
\end{tabular}}
\label{3mformer-msr}
\end{center}
\end{table}

\subsection{\leicr{Model Complexity}}

\leicr{Table~\ref{tab:complex-3mformer} shows the number of model parameters/FLOPs and NTU-60 accuracy. Our cost is moderate. 2S-AGCN (37.22 GFLOPs \& 3.45M param.) yields 89.4\% accuracy. Our `3rd-order' uses 35.5 GFLOPs \& 2.07M param. which is 2 GFLOPs \& 1.37M param. less, yet we outperform 2S-AGC by \textbf{1.9\%}. NAS-GCN uses 40.4 GFLOPs/2.2M param. more compared to our 3Mformer:  we beat NAS-GCN by \textbf{4.4\%}.}

\begin{table}[t!]
\caption{A comparison of the number of model parameters and FLOPs on NTU-60.}
\vspace{-0.5cm}
\begin{center}
\resizebox{\linewidth}{!}{\begin{tabular}{l c c c c c c}
\toprule
 & \multirow{2}{*}{ST-GCN} & \multirow{2}{*}{2S-AGCN} & \multirow{2}{*}{NAS-GCN} & 2rd-order & 3rd-order & 3Mformer\\
  &  & & & only ({\it ours}) & only ({\it ours}) & ({\it ours})\\
\midrule
Params (M) & 3.14 & 3.45 & 6.57 & 1.15 & 2.07 & 4.37\\
FLOPs (G) & 16.36 & 37.22 & 108.82 & 6.54 & 35.53 & 58.45\\
Acc. (\%) & 81.5 & 88.5 & 89.4 & 83.0 & 91.3 & {\bf 94.8}\\
\bottomrule
\vspace{-1.0cm}
\end{tabular}}
\label{tab:complex-3mformer}
\end{center}
\end{table}

\subsection{\leicr{Limitation and Future Work}}

\leicr{Despite the high accuracy of our model, there are still some limitations. Firstly, as we use $r$ branches of HoT, the number of parameters and computational cost are higher than existing methods. However, our method with single branch, \eg, 3rd-order HoT only, still achieves very competitive results compared to existing graph-, transformer- and hypergraph-based models for the same parameter scale on 3 benchmarks.
Secondly, in this work, we only use HoT block to encode the temporal block feature representations. The more efficient way is to redesign HoT block so that it is able to encode both short-term and long-term spatio-temporal features to simplify the backbone encoder, \ie, without the need of MLP unit. Note that the design of our 3Mformer is independent of the backbone encoder. Our 3Mformer is especially suitable for tensorial data, \eg, higher-order feature representations.
Our future work will focus on applying our Multi-order Multi-mode Transformer (3Mformer) to other computer vision tasks with tensorial data.
}

\end{document}